\documentclass{article}
\pdfoutput=1
\usepackage{ijcai22}

\usepackage{times}
\usepackage{soul}
\usepackage{url}
\usepackage[hidelinks]{hyperref}
\usepackage[utf8]{inputenc}
\usepackage[small]{caption}
\usepackage{graphicx}
\usepackage{subfigure}
\usepackage{amsmath}
\usepackage{amsthm}
\usepackage{booktabs}
\usepackage{algorithm}
\usepackage{algorithmic}
\usepackage{varwidth}
\title{Towards Explainability in NLP: Analyzing and Calculating Word Saliency through Word Properties}

\author{
Jialiang Dong$^1$
\and
Zhitao Guan$^1$\and
Longfei Wu$^2$\and
Zijian Zhang$^3$\and
Xiaojiang Du$^4$
\affiliations
$^1$North China Electric Power University\\
$^2$Fayetteville State University\\
$^3$Beijing Institute of Technology\\
$^4$Stevens Institute of Technology
}

\begin{document}

\maketitle

\begin{abstract}
  The wide use of black-box models in natural language processing brings great challenges to the understanding of the decision basis, the trustworthiness of the prediction results, and the improvement of the model performance. The words in text samples have properties that reflect their semantics and contextual information, such as the part of speech, the position, etc. These properties may have certain relationships with the word saliency, which is of great help for studying the explainability of the model predictions. In this paper, we explore the relationships between the word saliency and the word properties. According to the analysis results, we further establish a mapping model, Seq2Saliency, from the words in a text sample and their properties to the saliency values based on the idea of sequence tagging. In addition, we establish a new dataset called PrSalM, which contains each word in the text samples, the word properties, and the word saliency values. The experimental evaluations are conducted to analyze the saliency of words with different properties. The effectiveness of the Seq2Saliency model is verified.
\end{abstract}

\section{Introduction}

Deep Neural Networks (DNNs) have achieved great success in Natural Language Processing (NLP) tasks. However, the DNNs based models are usually black-box models in actual scenarios and it is hard to understand the internal working mechanisms of these models. Analyzing the relationship between each part of the text samples and the prediction results of the model can help to better understand the basis for prediction, so as to unveil the potential biases and explain the model predictions~\cite{arrieta2020explainable,doshi2017towards}. Furthermore, understanding the principles of  the model is the basis for improving model performance~\cite{miller2019explanation}.

Analyzing the saliency distribution of natural language samples is a great challenge. In NLP tasks, samples are generally composed of dozens or even hundreds of words, and the semantics are highly based on the sequence information. The saliency of distinct words depends not only on their contents, but also on their contexts. Most of the existing black-box saliency calculation methods in the field of NLP rely on comparing the prediction results of the original sample with its approximate representations. These methods usually require a large number of visits to the model for getting the prediction results, which is hard to realize in actual scenarios and cannot make full use of the rich features contained in the samples.

\begin{figure}[t]
	\centering
	\includegraphics[width=\columnwidth]{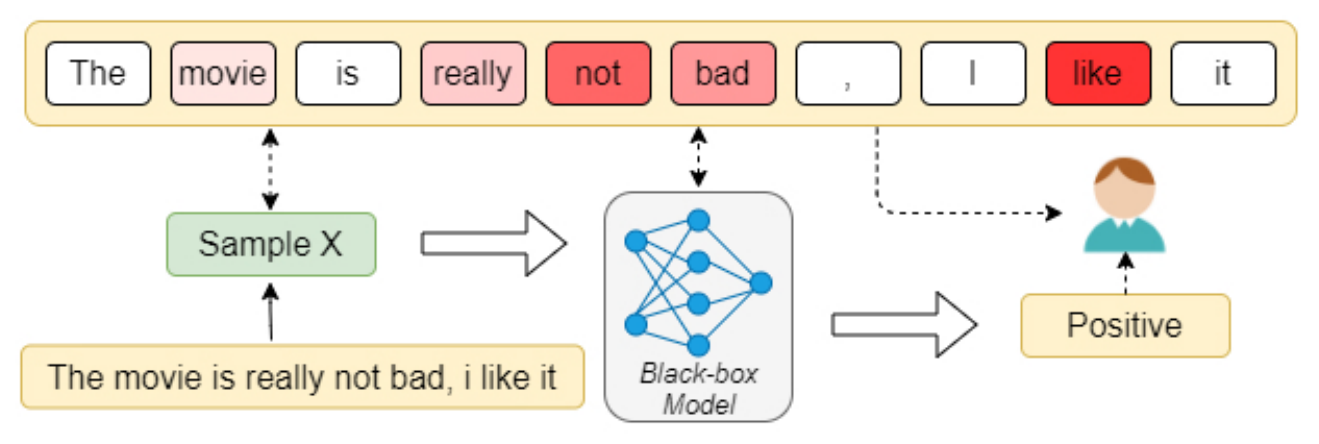}
	\caption{An example of word saliency heat map.}
\end{figure}

The words in text samples have properties which reflect the semantics and contexts, such as the part of speech (POS), position, and so on. These properties can reflect the importance of the words to the text sample. For example, the distribution of word saliency in text samples is related to specific tasks, while nouns, adjectives, and verbs are usually regarded as content words that carry meanings~\cite{zang-etal-2020-word}. Understanding the relationships between the various properties of words and the model output can help to establish the mapping between the samples and their saliency distribution, and to analyze the contribution of each part of the samples to the outputs of the model. 

In this paper, we explore a new way to explain NLP model predictions by analyzing and utilizing word properties in text samples. We analyze the relationships between the word properties and the saliency values. Also, we employ the idea of sequence tagging to train an end-to-end model to calculate the word saliency values. The main contributions of this paper are summarized as follows.

\begin{itemize}
	\item We explore the relations between the word properties and the word saliency in natural language samples through quantitative analysis.
	\item Based on the obtained mapping data, we employ the idea of sequence tagging to train an end-to-end word-saliency mapping model, which we call Seq2Saliency, to calculate the word saliency values in NLP tasks.
	\item We construct a new dataset PrSalM for the mapping between the words in text samples (as well as the word properties) and saliency values. To the best of our knowledge, PrSalM is the first dataset to take word saliency value as word tagging information, which can be used for future research on word property analysis and model explainability.
\end{itemize}

\section{Preliminary}
\subsection{Related Works}

\begin{table}[t]
	\centering
	\begin{tabular}{ccccc}
		\toprule
		Dataset & Train.Size & Test.Size & Avg.Len & Task \\
		\midrule
		SST-2     & 20K  & 1K & 17 & Class.     \\
		AG's News & 20K  & 1K & 43 & Class.     \\
		SNLI      & 200K & 1K & 8  & Inf.       \\
		\bottomrule
	\end{tabular}
	\caption{Details of Datasets}
	\label{tab:booktabs}
\end{table}

\begin{table}[t]
	\centering
	\begin{tabular}{ccc}
		\toprule
		Model & Pretrained Version & Tokenization \\
		\midrule
		BERT       & ``bert-base-uncased"       & WordPiece \\
		RoBERTa    & ``roberta-base"            & BPE     \\
		DistilBERT & ``distilbert-base-uncased" & WordPiece \\
		\bottomrule
	\end{tabular}
	\caption{Details of Pretrained Models}
	\label{tab:booktabs}
\end{table}

In terms of explaining the predictions of the classifiers, Ribeiro et al.~\cite{ribeiro2016should} proposed to learn an interpretable model which is locally around the prediction. For deep learning tasks, when studying the contribution of each part, it is an intuitive idea and a common method to use the gradient of output for expressing the saliency of each input part. But in NLP tasks, the input words are mostly expressed in the form of vectors, which makes it difficult to calculate the saliency values in word granularity. INTGRAD, presented in~\cite{sundararajan2017axiomatic}, is immune to the above problem when used in NLP tasks because the calculation is realized by a difference to the baseline value. Shrikumar et al. ~\cite{shrikumar2017learning} proposed to backpropagate the contributions of all neurons in the DNN model to each feature of the input, and analyze the saliency of different features. Chen et al.~\cite{chen2018learning} introduced a method called instancewise feature selection, which trains a feature selector that can maximize the mutual information between the chosen features and the output, to extract a subset of the most informative features in samples. For text samples, Ribeiro et al.~\cite{ribeiro2018anchors} proposed to replace words with UNK, to analyze the contribution of different words in the samples to the output results of the model. Besides, the unified framework SHAP~\cite{lundberg2017advances} approximates the Shapley value of the model conditional expectation function by combining the existing calculation methods through reasonable assumptions. In addition, there are some other schemes~\cite{li2016understanding,cheng-etal-2016-long} based on attention mechanism to study the explainability of the model’s outputs, while the significance of the attention value to saliency value remains to be discussed~\cite{jain-wallace-2019-attention,wiegreffe-pinter-2019-attention}.

An example of word saliency visualization expressed by the saliency heat map is shown in Figure 1. The sample corresponds to the task of emotion analysis in movie reviews. Different shades of highlight indicate the intensity of emotion polarity, which are reflected in the saliency values. Even if the model can predict samples with high accuracy, the lack of explainability makes the prediction results unreliable. The saliency analysis can help to judge whether or not the prediction basis of the model is consistent with human beings, so as to evaluate the reliability of the results.

\subsection{SHAP Framework}
SHAP\footnote{https://github.com/slendberg/shap} is a framework that unifies several saliency calculation methods, such as Shapley values~\cite{shapley201617}, LIME, and DEEPLIFT. It assigns each feature a saliency value for a specific prediction, which can be used for both model-agnostic approximations and model-specific approximations. It is based on the best Shapley value in game theory.

\newcommand{\mysize}{5.2cm}
\begin{figure*}[t]
	\centering
	\subfigure[SST-2]{
		\includegraphics[width=\mysize]{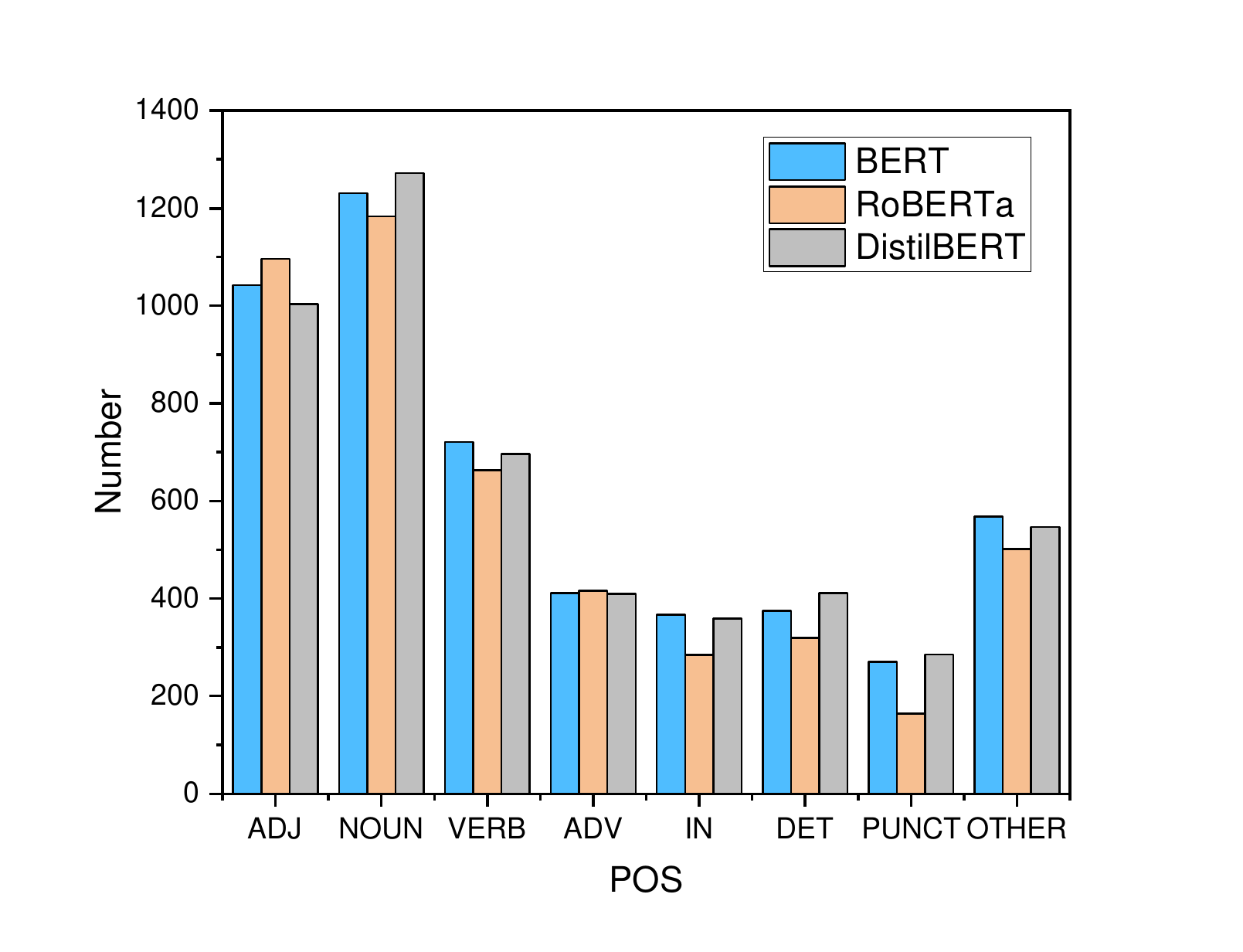}
	}
	\qquad
	\subfigure[AG's News]{
		\includegraphics[width=\mysize]{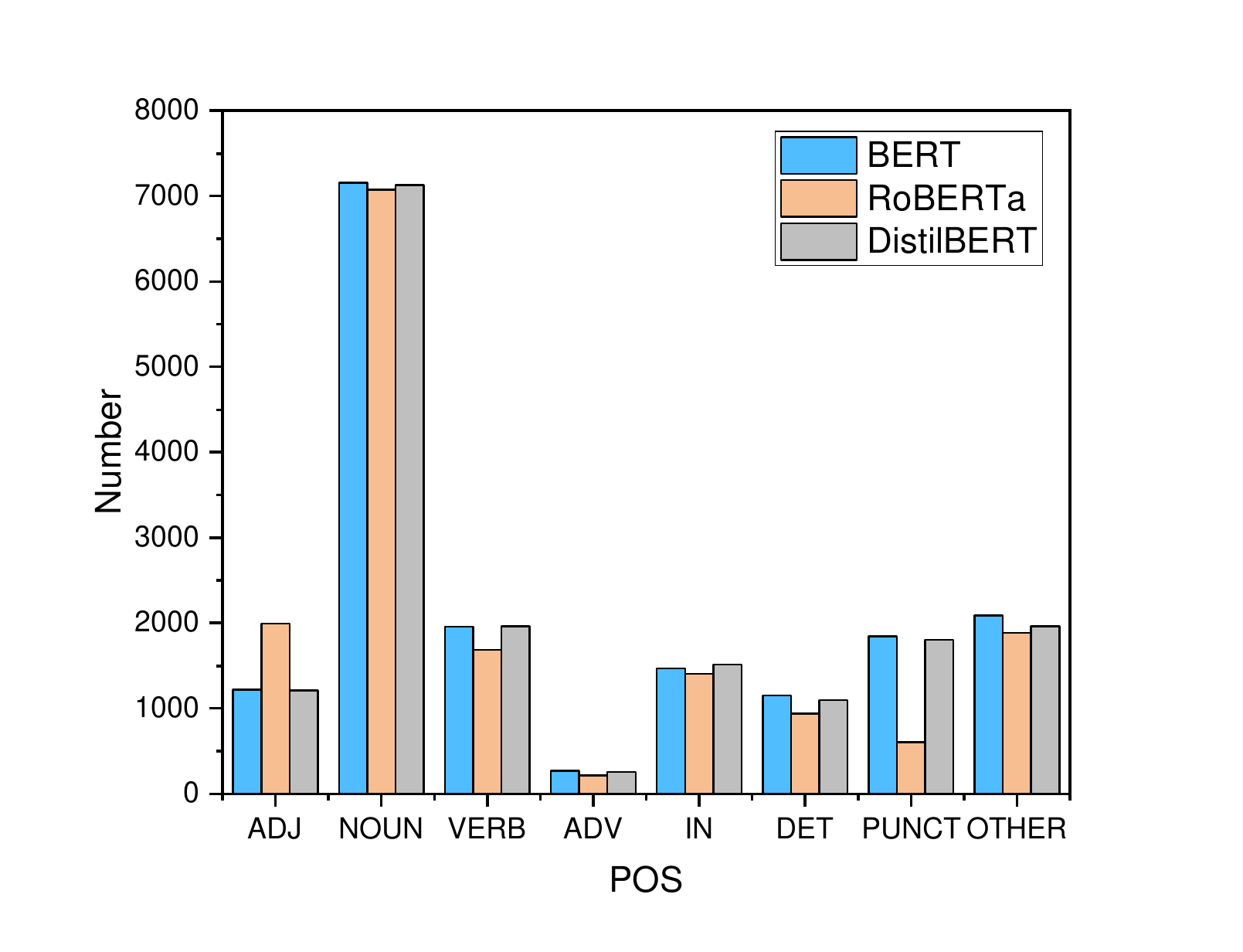}
	}
	\qquad
	\subfigure[SNLI]{
		\includegraphics[width=\mysize]{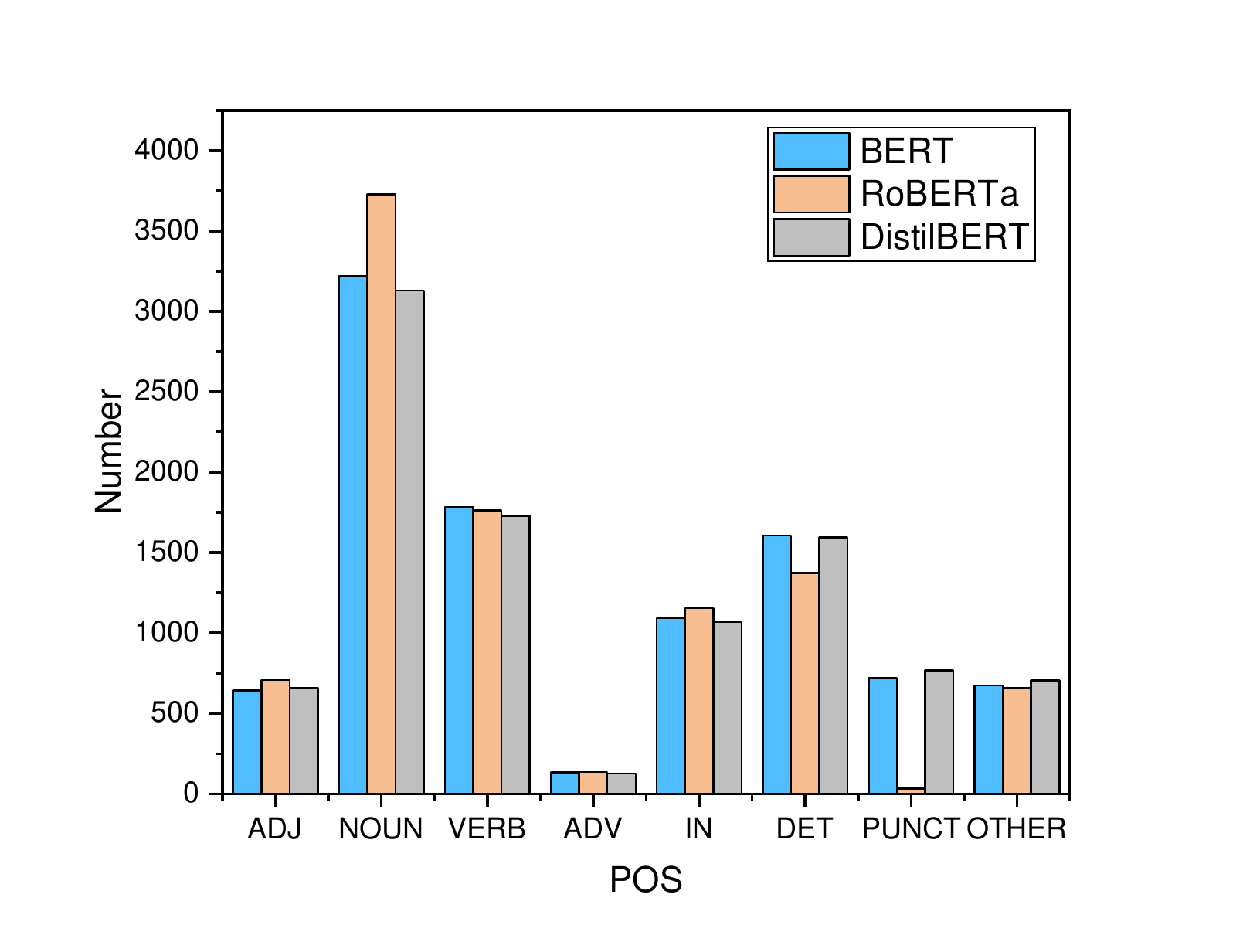}
	}
\caption{The number of distinct POS of words in the top 30\% saliency values. The rectangles with different colors correspond to the results calculated based on different pretrained models.}
\end{figure*}

\begin{figure*}[t]
	\centering
	\subfigure[SST-2]{
		\includegraphics[width=\mysize]{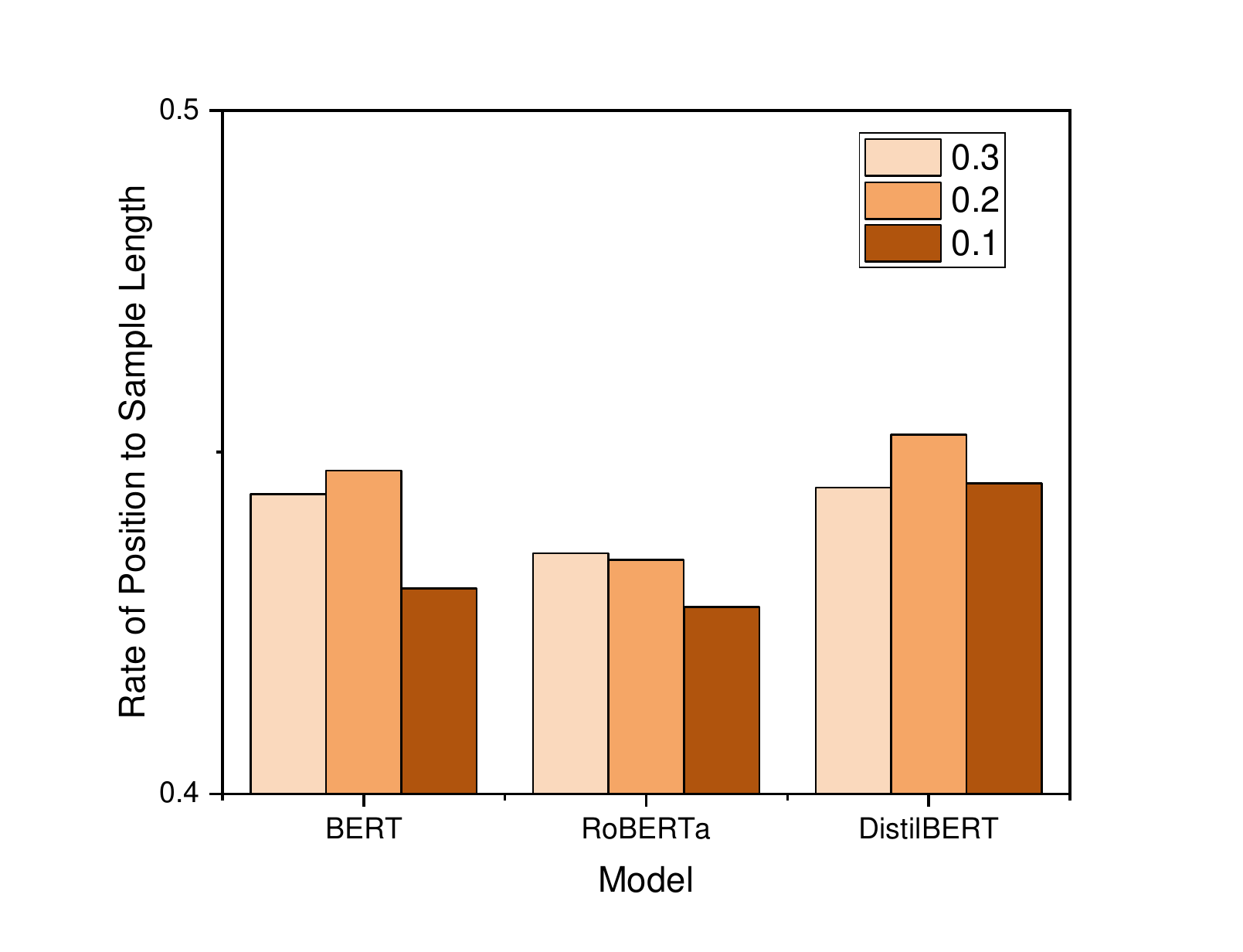}
	}
	\qquad
	\subfigure[AG's News]{
		\includegraphics[width=\mysize]{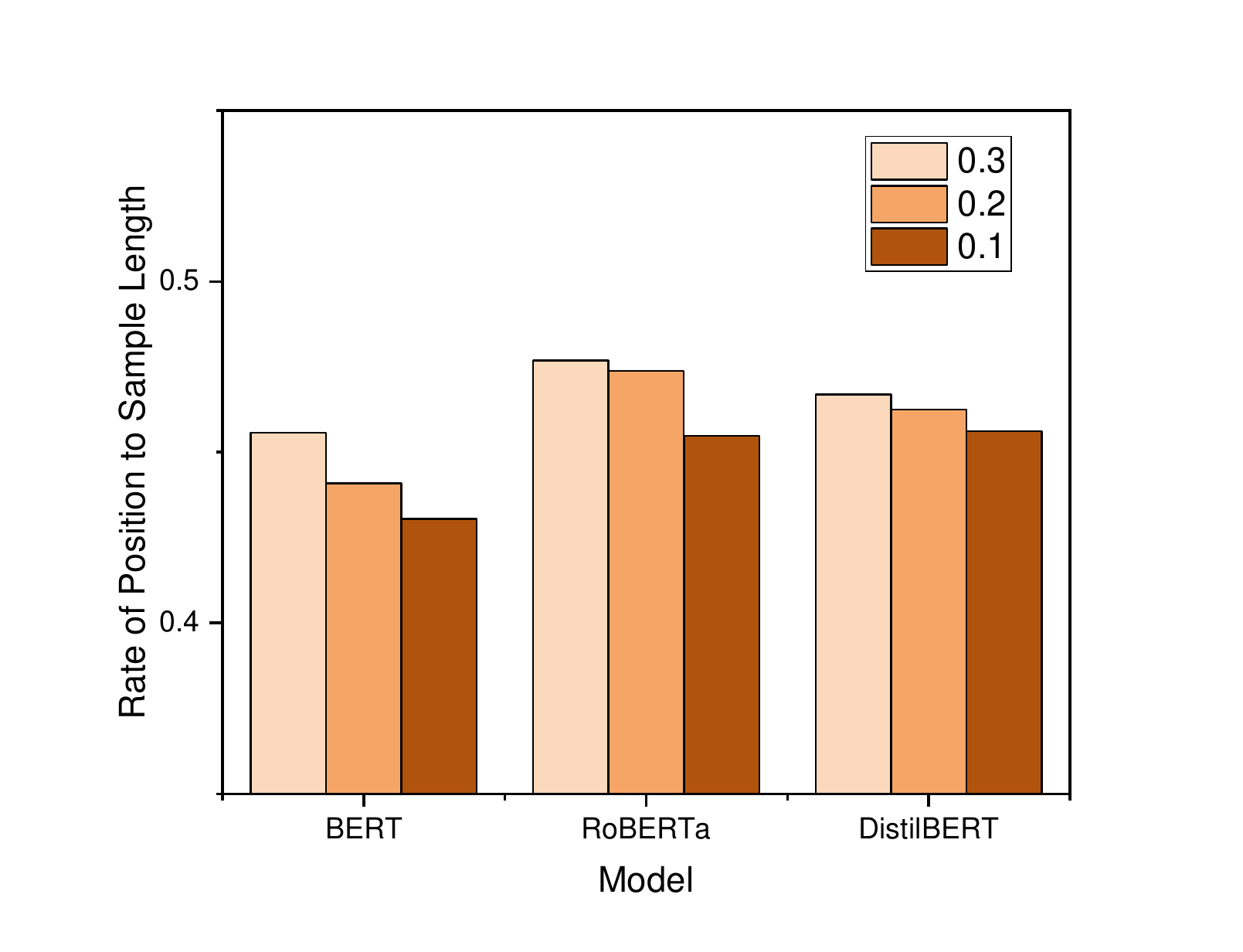}
	}
	\qquad
	\subfigure[SNLI]{
		\includegraphics[width=\mysize]{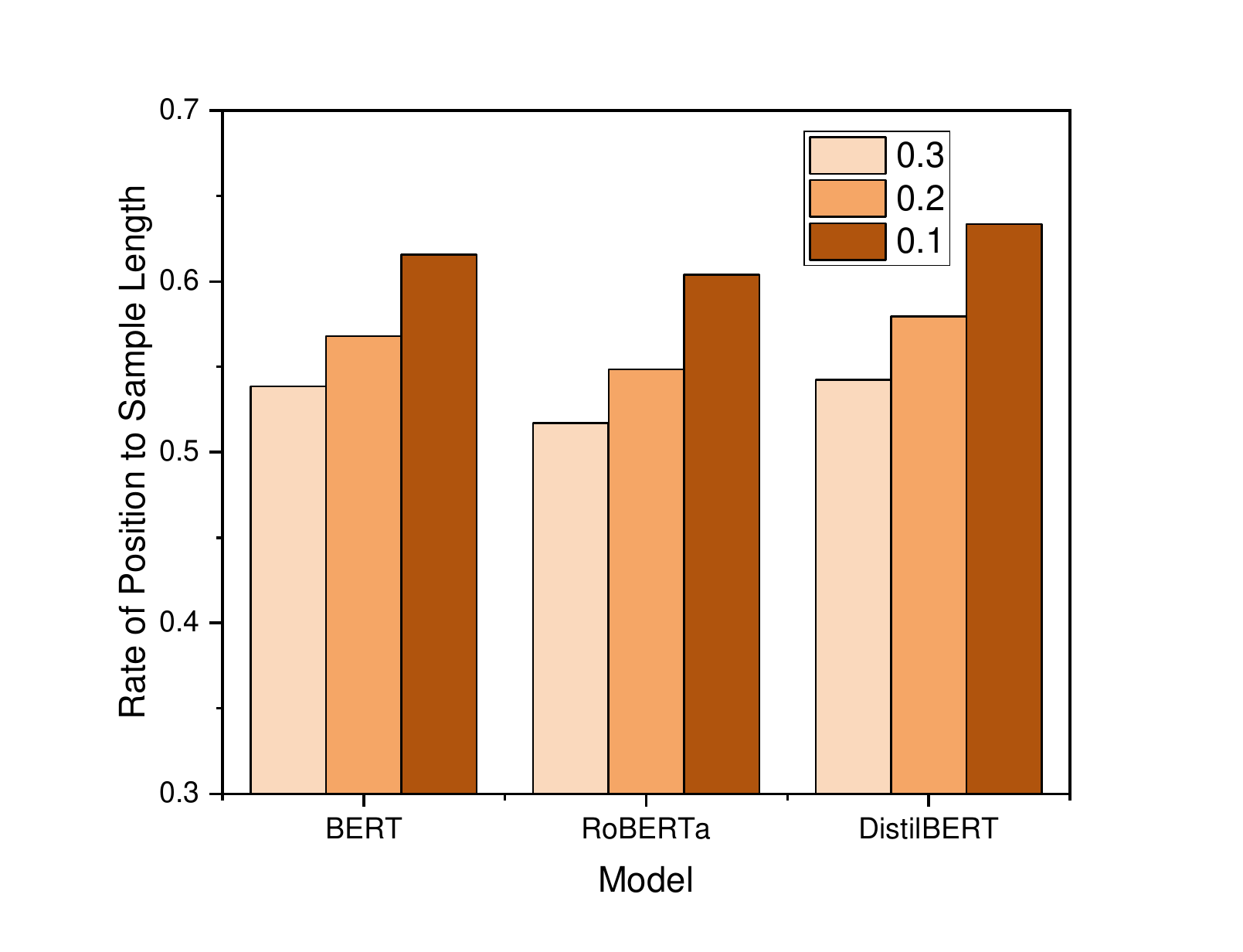}
	}
	\caption{The average value of the ratio of word position to sample length in the top theta of saliency value. Here the orange rectangles with different lightness represent the statistical results corresponding to different theta.}
\end{figure*}

\begin{figure*}[t]
	\centering
	\subfigure[SST-2]{
		\includegraphics[width=\mysize]{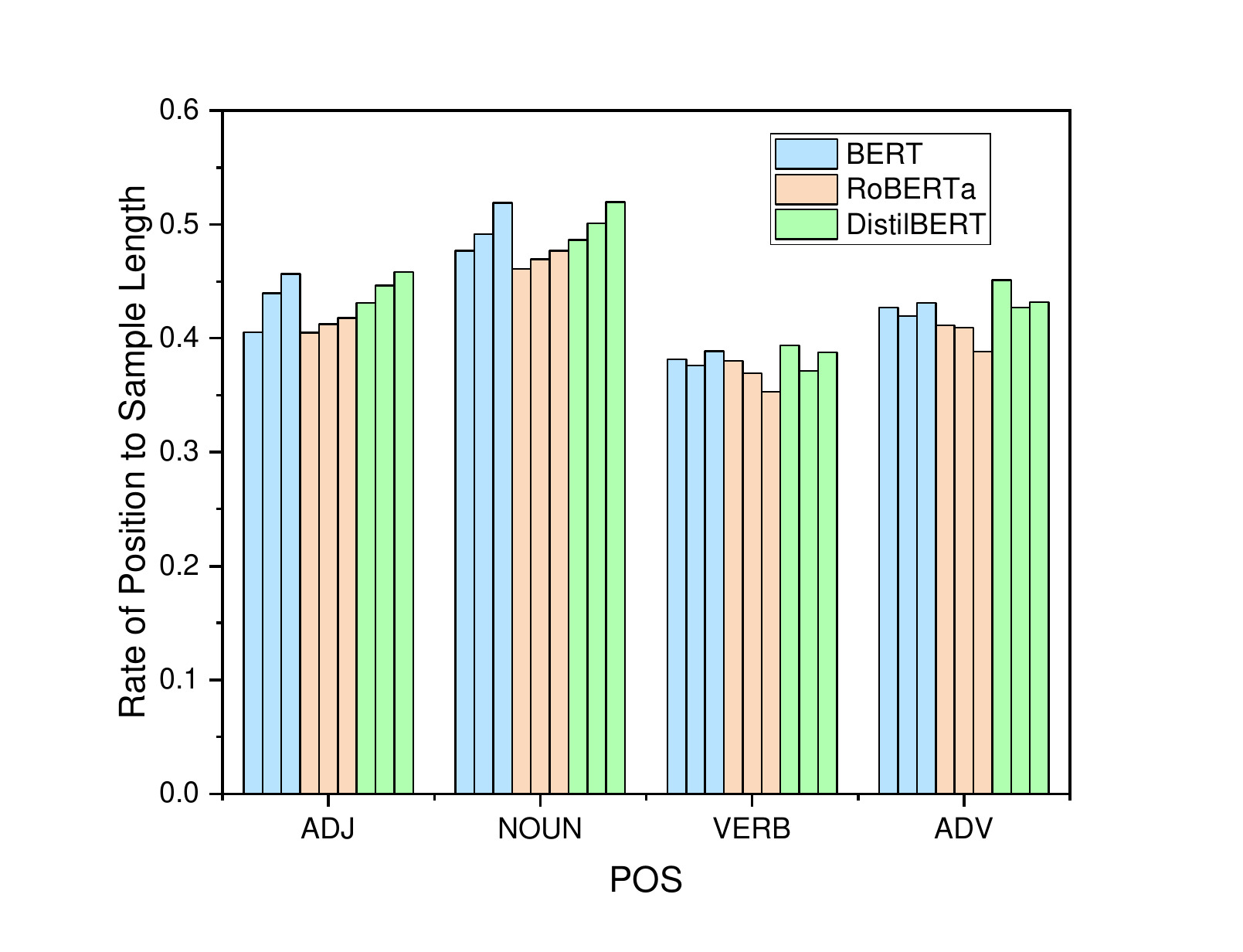}
	}
	\qquad
	\subfigure[AG's News]{
		\includegraphics[width=\mysize]{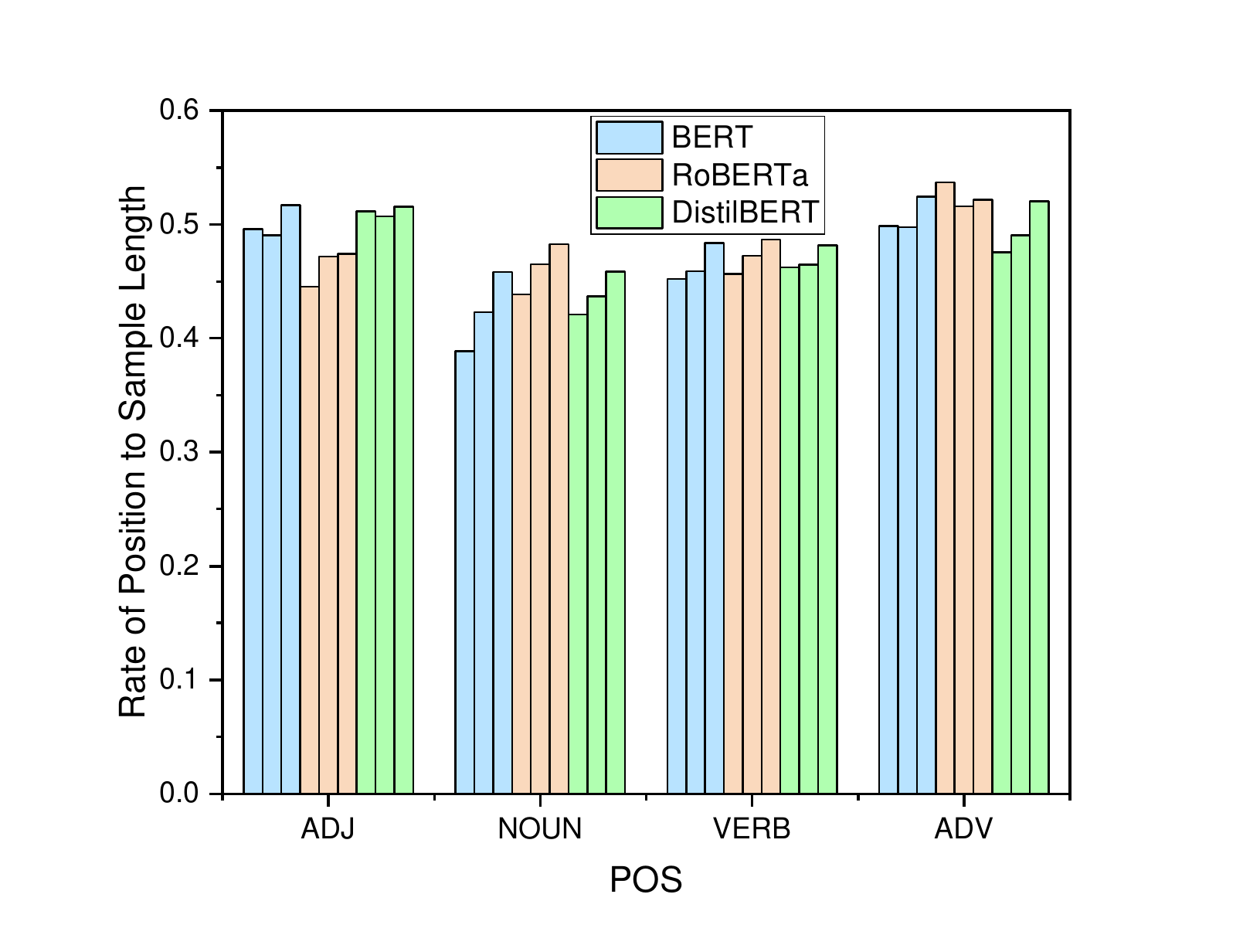}
	}
	\qquad
	\subfigure[SNLI]{
		\includegraphics[width=\mysize]{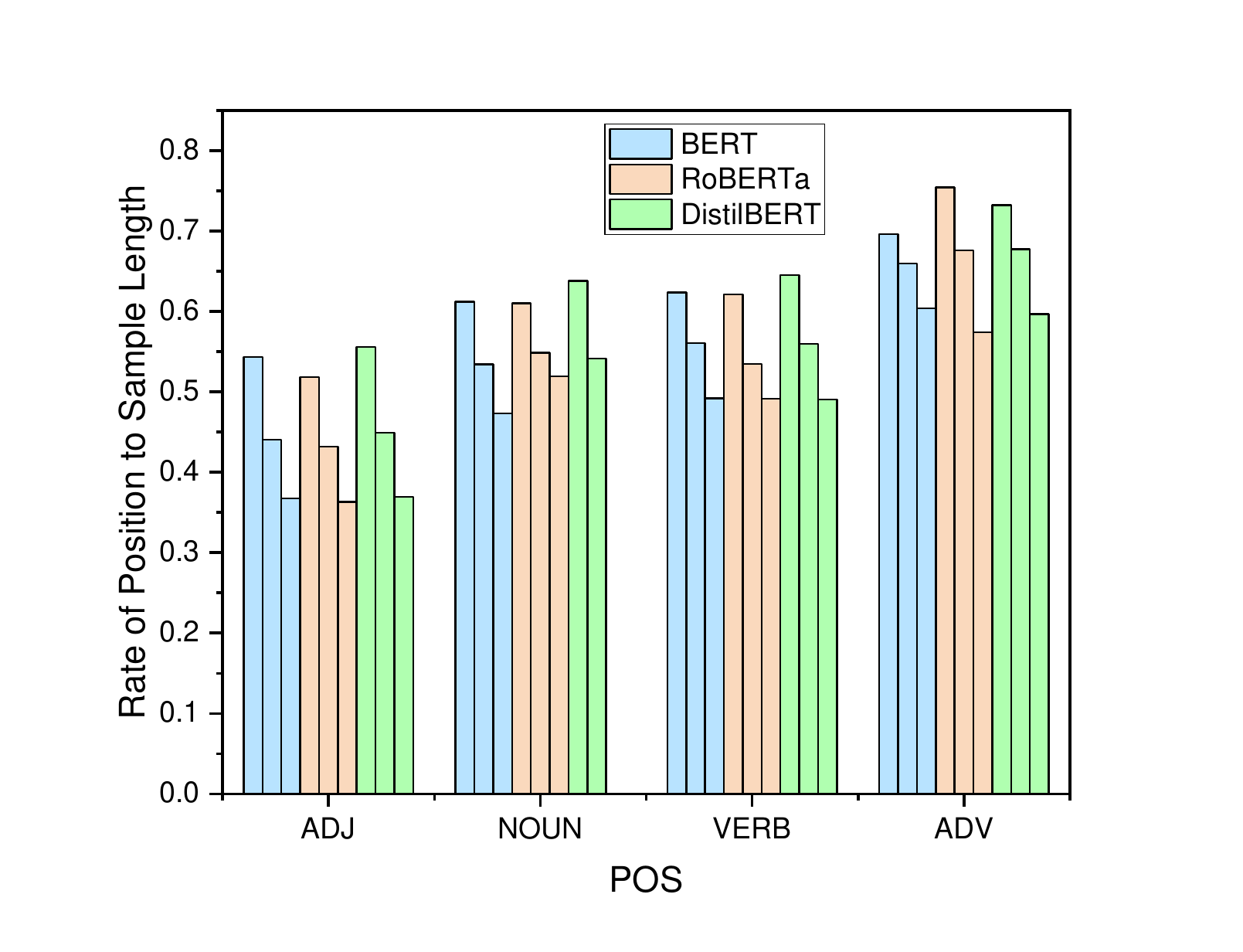}
	}
	\caption{The average value of the ratio of the word position with specific POS to the sample length in the top theta of saliency value. For each group of three rectangles in the same color, from left to right the rectangles represent the position distribution in the situations that the theta value is 0.1, the theta value is 0.3, and all the words of the POS are counted, respectively.}
\end{figure*}

\begin{figure*}[t]
	\centering
	\subfigure[SST-2]{
		\includegraphics[width=\mysize]{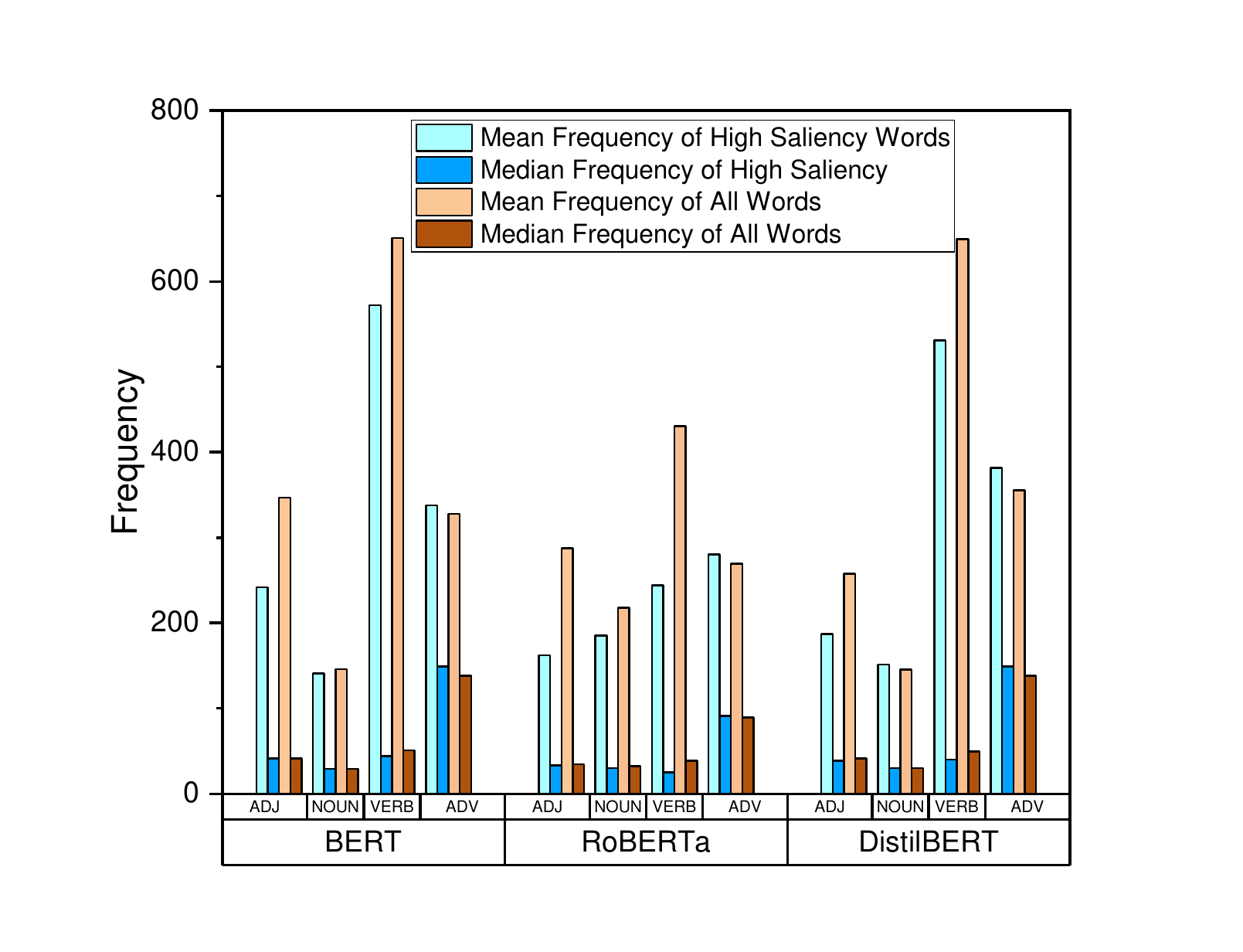}
	}
	\qquad
	\subfigure[AG's News]{
		\includegraphics[width=\mysize]{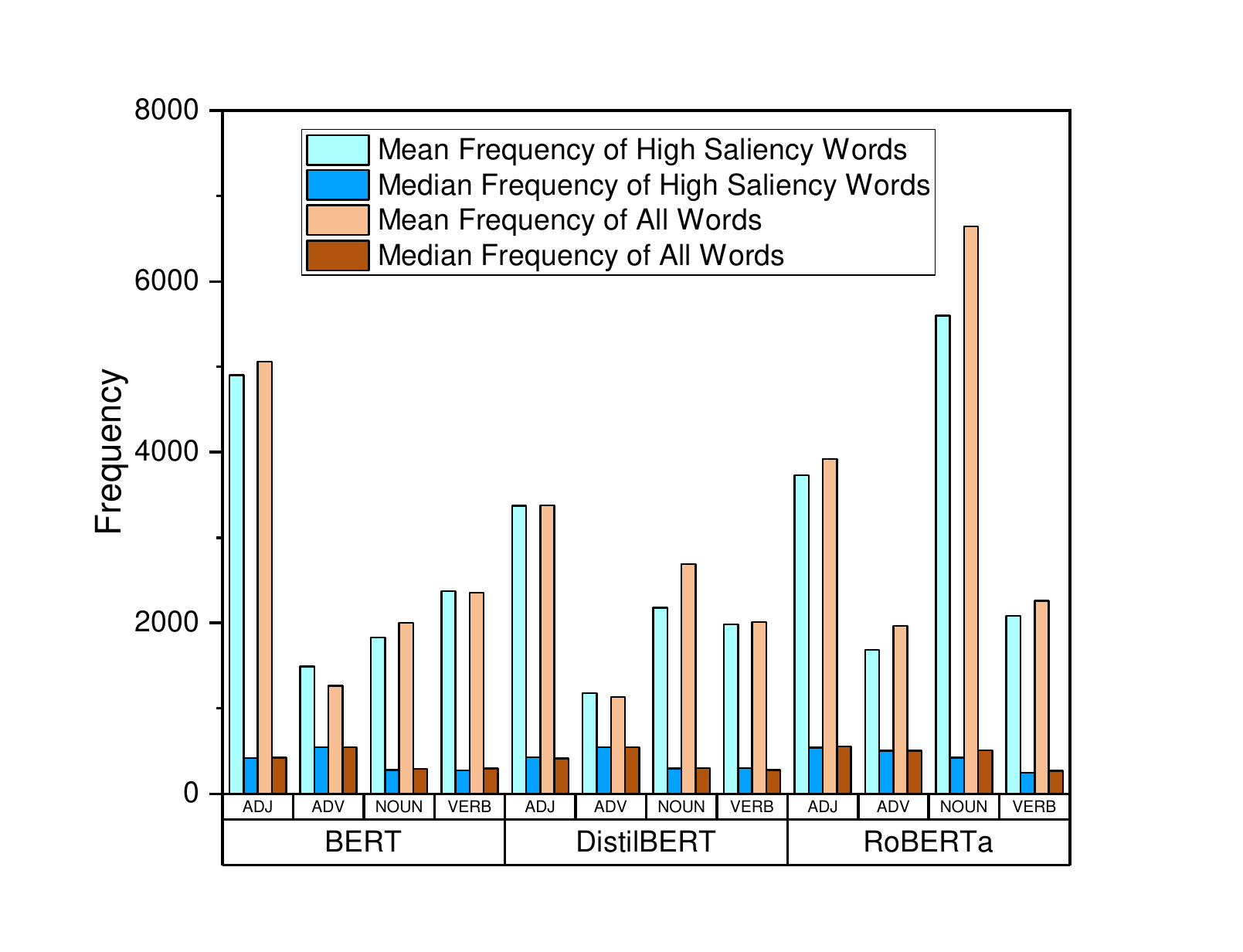}
	}
	\qquad
	\subfigure[SNLI]{
		\includegraphics[width=\mysize]{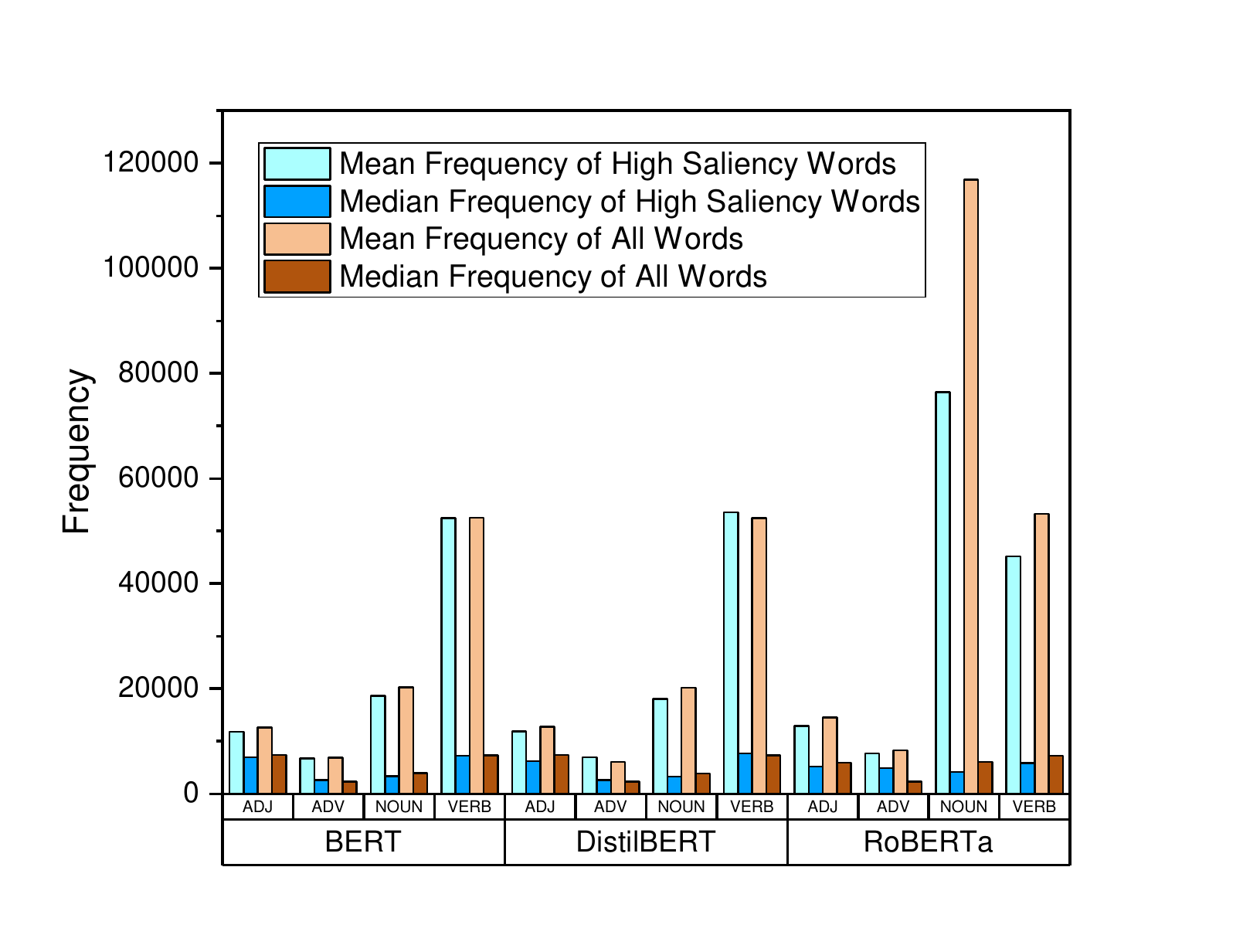}
	}
	\caption{The statistics of word frequency distribution. The blue rectangles represent the frequency statistics of words with high saliency values (top 30\%), the orange rectangles represent the frequency statistics of all words. The light color rectangles represent the statistics of the mean word frequency, and the dark color rectangles represent the statistics of the median word frequency.}
\end{figure*}

SHAP values are difficult to compute. The authors proposed to combine insights from other additive feature attribution methods. The approximate calculation of the SHAP value is realized by reasonable assumptions and simplifications. When taking SHAP to analyze NLP tasks, it is realized by adding coalitional rules to Shapley values.

Here, we use the SHAP Framework to calculate the word saliency values, which are used to analyze the relationships between the word properties and the word saliency, and to train a new saliency calculation model.

\subsection{Properties, Datasets, and Models}
In human intuition, words with different properties have different contributions to the final semantic representation and the prediction results. For DNN models, this intuitive rule is worth studying.

Here we choose the important word properties including the POS, the position, and the frequency of words to study. The word frequency here refers to the frequency of words appearing in the training dataset.

To make the analysis results more convincing, we choose three datasets, SST-2, AG’s News, and SNLI, for the classification and inference tasks in our experiments. The detailed information of these datasets is shown in Table 1. The terms "Class." and "Inf." stand for the classification task and the inference task, respectively.

Besides, we choose three different pre-trained models for task implementation, including BERT~\cite{devlin-etal-2019-bert}, RoBERTa~\cite{liu2019roberta}, and DistilBERT~\cite{sanh2019distilbert}. We use the pre-trained models in transformers library\footnote{https://github.com/huggingface/transformers} and fine-tune each model with the same training datasets.

\section{Property-Saliency Relationship Analysis}
In this section, we adopt the existing saliency calculation method SHAP to calculate the word saliency values. At the same time, we obtain word properties including the POS, position, and frequency, and analyze the relationships between each of these word properties and the saliency values. Note that different word tokenization methods will get different word sequences. In order to carry out effective analysis, we use the respective word tokenization methods that are consistent with the pre-trained models.

In particular, we analyze the relationships by calculating the distribution of word properties in different saliency value intervals. The quantitative calculation results are obtained based on 1000 test samples.

\subsection{POS-Saliency Relationship Analysis}
POS is an important property that has been paid a lot of attention in the study of language model learning content analysis~\cite{qian2016analyzing,dalvi2019one}. We finetune the pre-trained TokenClassification models\footnote{BertForTokenClassification (‘bert-base-uncased’) and RoBertaForTokenClassification (‘roberta-base’) pretrained models are used for POS tagging} in the transformers library with conll2003 dataset\footnote{https://huggingface.co/datasets/conll2003} as the POS tagging models. For the 1000 test samples, we calculate the word saliency values and analyze the POS distribution of the words with high saliency values.

As shown in Figure 2, the POS listed here are those with larger number of words. Each group of rectangles with different colors represent the mean number of words whose saliency values are in the top 30\%, with a specific POS under different models. The results show that no matter which dataset or model is used, nouns, adjectives, and verbs account for a relatively high proportion of words with high saliency values. Meanwhile, for distinct datasets and models, the saliency distribution may change greatly. For instance, when the dataset is SST-2, adverbs account for an obviously higher proportion of the words with high saliency, compared to the situations in which the dataset AG's News or SNLI is used. However, for a specific task, there is still a clear correlation between the POS of a word and the saliency value of the word.

\subsection{Position-Saliency Relationship Analysis}
The semantics of natural language samples are highly dependent on the sequence information, so that the word position in the text sample is also an important property~\cite{adi2017fine}. Here we study the relationship between the word position and the word saliency. For each sample, we calculate the ratio of the word position to the number of words in the whole sample (sample length) and analyze the relationship between this ratio and the saliency value.

In Figure 3, the orange rectangles with different lightness represent the average position distribution of the words with the top theta saliency, respectively. For classification tasks on the datasets SST-2 and AG’s News, the average position of words with higher saliency values is usually distributed evenly and a bit closer to the front when the theta value decreases (from 0.3 to 0.1). However, when the task is an inference task on the SNLI dataset, the average position of words with higher saliency in the sample goes backward with the decrease of theta. Therefore, we can conclude that for classification tasks, the words with high saliency are evenly distributed and slightly biased towards the first half, while for inference tasks, the words with higher saliency are more likely to be distributed in the second half.

In addition to analyzing the relationship between the position distribution and the saliency for all words, we also calculate the position distribution of words with high saliency for specific POS. As shown in Figure 4, we choose the POS (noun, verb, adjective, adverb) with relatively high average saliency values to analyze. For each group of three rectangles with the same color, the rectangles represent the position distribution in the situations that the theta value is 0.1, the theta value is 0.3 and all the words of the POS are counted, respectively. The results show that the distribution characteristics of the individual POS in different types of tasks are a bit different, but they still conform to the rules obtained from the analysis on all the words.

\subsection{Frequency-Saliency Relationship Analysis}
The word frequency we studied refers to the number of times a word appears in the training dataset. It is generally assumed that the more the words appear in the training dataset, the better the model can master them~\cite{mozes-etal-2021-frequency}. For all the words appearing in the training dataset, we count the frequency to analyze the relationship between the word frequency and the saliency value.

In Figure 5, for each POS, we count the mean frequency and the median frequency of all words of this POS as well as the words with high saliency values (saliency values in the top 30\%). Since the word frequency can vary greatly in the training dataset, the combined use of the mean frequency and the median frequency can better reflect the distribution of word frequency. Here, the blue and orange rectangles represent the frequency statistics of high saliency words and all words, respectively. The light and dark colors represent the mean frequency and median frequency, respectively. As shown in Figure 5, the mean frequency of words with high saliency is lower than that of all words in many cases. However, considering that the results based on different datasets and models are not consistent, and that there is no obvious difference in the median frequency, we conclude that the relationship between the word frequency and saliency value is not certain.

\begin{figure}[t]
	\centering
	\includegraphics[width=8.3cm]{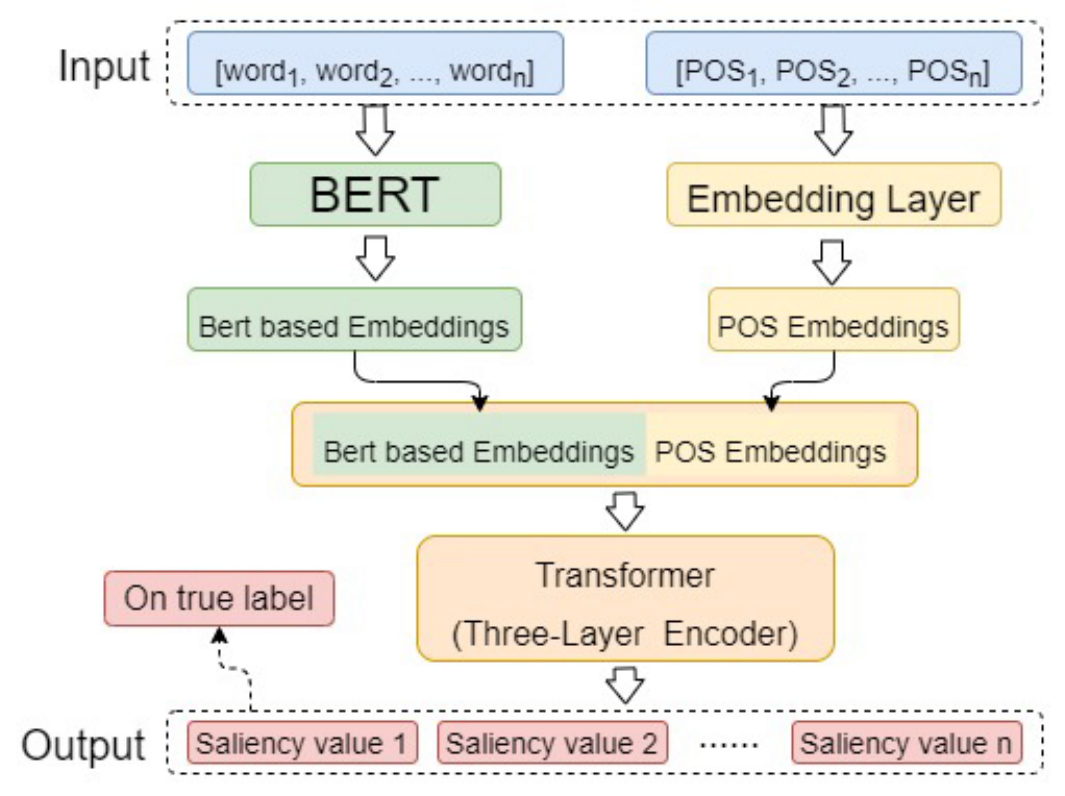}
	\caption{The structure of Seq2Saliency model.}
\end{figure}

\section{Seq2Saliency Mapping}
\subsection{Properties-Saliency Mapping Dataset}
In the section of property-saliency relationship analysis, we calculated the word properties and the saliency value. Based on the original samples, the calculated properties and the saliency values, we further construct a new dataset named Properties-Saliency Mapping (PrSalM) dataset. The dataset consists of 6 parts bulit on 3 datasets (SST-2, AG’s News, SNLI) with 2 tokenization methods (WordPiece~\cite{devlin-etal-2019-bert}, BPE~\cite{sennrich-etal-2016-neural}). The details of PrSalM are shown in Appendices.

\subsection{Seq2Saliency Model}
According to the results of property-saliency analysis, there are specific relationships between the word properties and the saliency. Based on this, we believe that the text sample itself contains abundant information that can reflect the saliency of each part. We use the idea of sequence tagging to fit the relationship between words and the saliency values, with the word itself and the word properties as the input information. The details are described below.

\begin{equation*}
	\small
	S(words, properties)=SaliencyValues\tag{1}
\end{equation*}

\begin{equation*}
	\small
	properties=(POS_i,position_i),i=1,2,...,n\tag{2}
\end{equation*}

\begin{equation*}
	\small
	Value_{norm}=\frac{Value-Value_{min}}{Value_{max}-Value_{min}}\tag{3}
\end{equation*}

In Equation (1), the “words” refers to a text sample which is composed of a sequence of words obtained by tokenizing the original text sample. The “properties” refers to a set of word properties with the same length as “words”. $S$ describes the function of the Seq2Saliency model, that is, to realize the mapping from words to saliency values. As shown in Equation (2), the “properties” of a word includes the POS and the position. In particular, we use Min-Max scaling to scale the saliency value of each sample. As described in Equation (3), given a sample, $Value$is the saliency value of a word in the sample, while $Value_{min}$ and $Value_{max}$ represent the minimum and the maximum saliency values in this sample, respectively.

\begin{equation*}
	\small
	Loss(S(X),Y) = \frac 1n \sum_{i=1}^n (Y_i-S(X_i))^2\tag{4}
\end{equation*}

\begin{figure*}[t]
	\centering
	\subfigure[SST-2]{
		\includegraphics[width=\mysize]{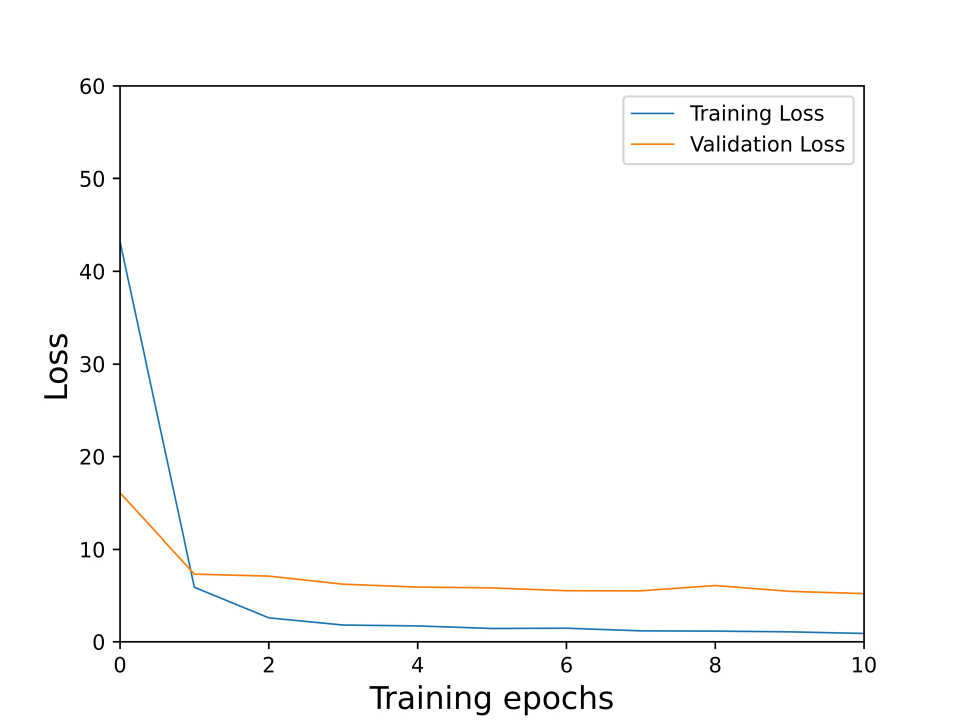}
	}
	\qquad
	\subfigure[AG's News]{
		\includegraphics[width=\mysize]{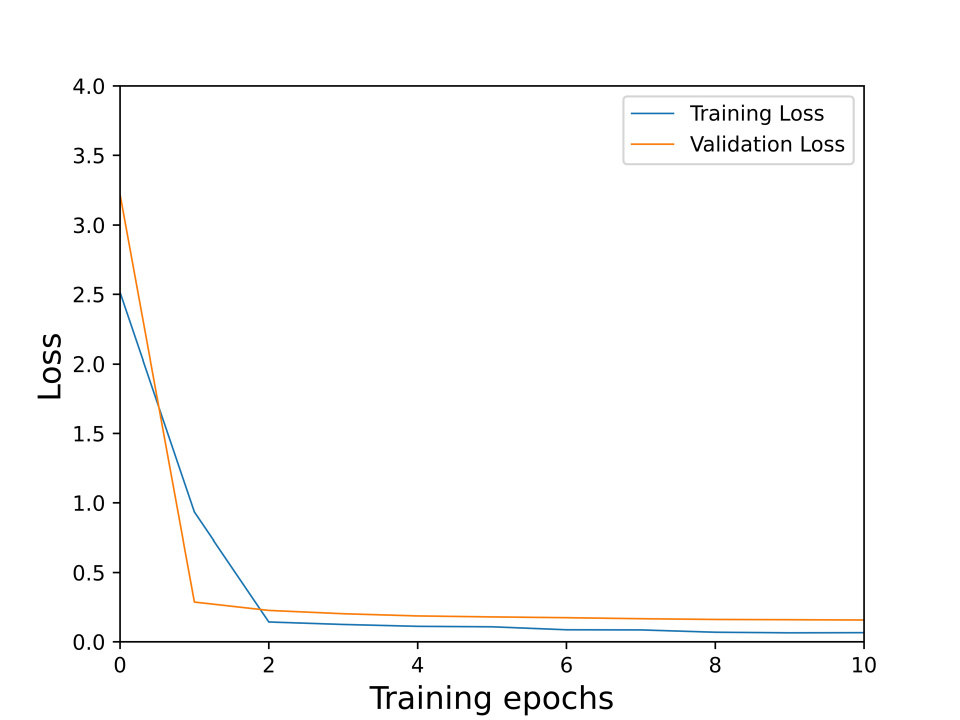}
	}
	\qquad
	\subfigure[SNLI]{
		\includegraphics[width=\mysize]{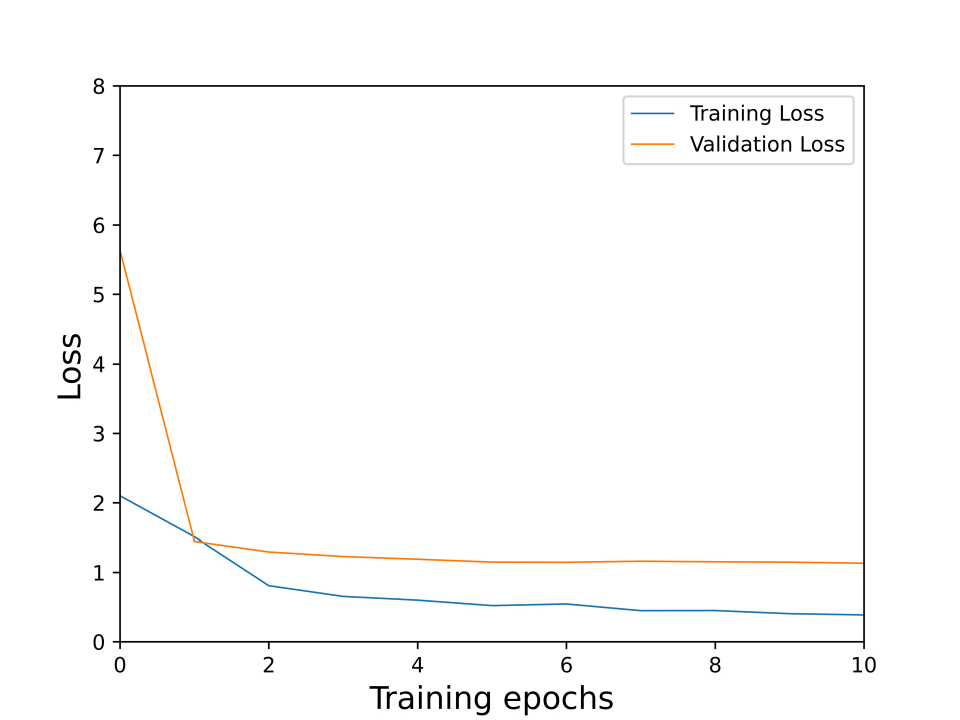}
	}
	\caption{The change of the loss values during the training process. Blue lines and orange lines represent training loss and validation loss, respectively.}
\end{figure*}

\begin{figure}[t]
	\centering
	\includegraphics[width=8.3cm]{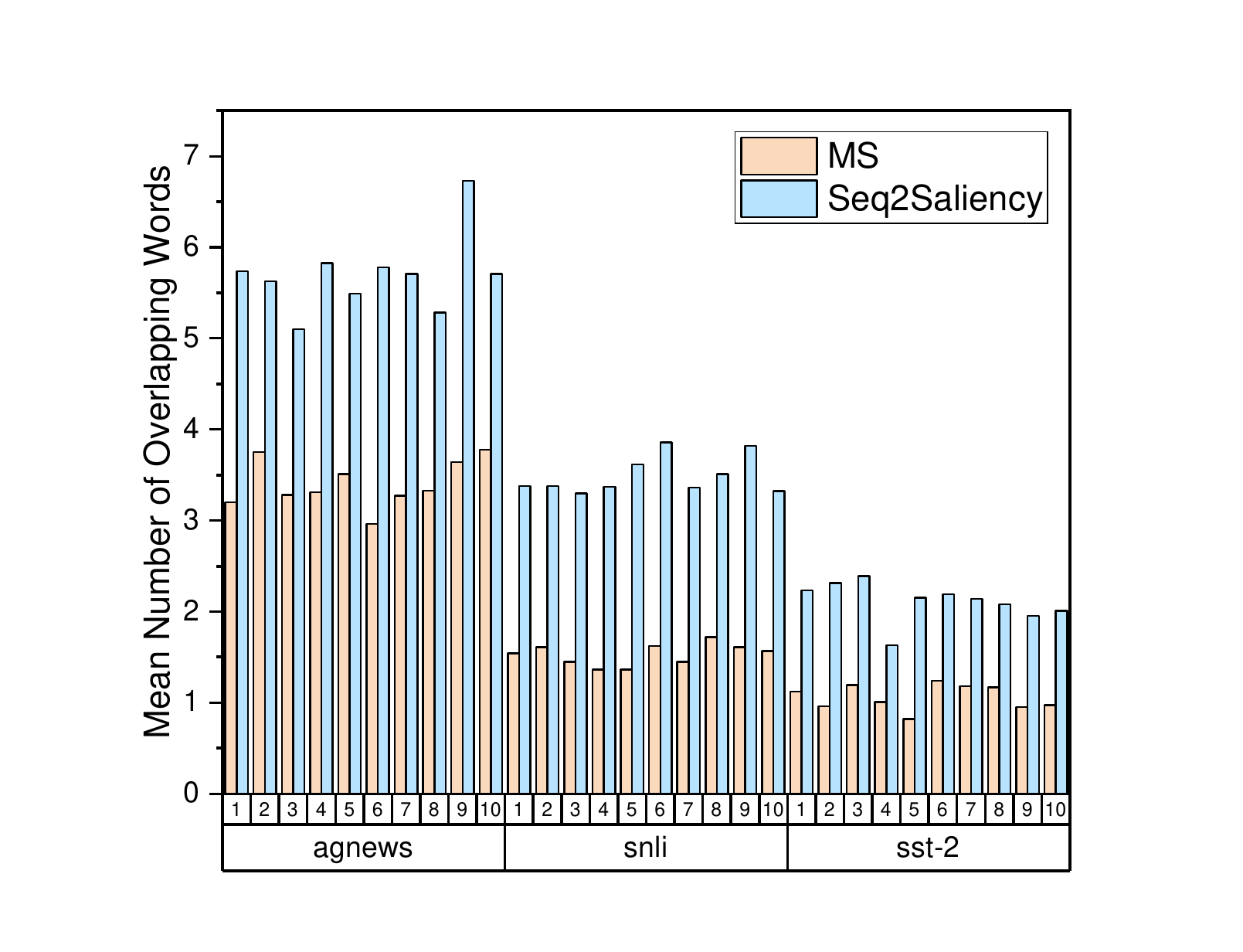}
	\caption{The mean value of overlapping words in top 30\% saliency values.}
\end{figure}

We train the Seq2Saliency model by using the words, the word POS, and the word positions as the input of the model, while the saliency values (calculated with the SHAP framework) are used as the supervision values. We take MSELoss as the loss function to train the model, as shown in Equation (4), in which the input is represented by X and the supervision information (saliency values) is Y. We use the pre-trained language model (BERT model) to extract the feature of the words and use the model’s output to represent the words. In the process of extracting word features, the position information of words is also involved in the calculation. A trainable embedding layer is used to encode the word POS. Finally, we adopt a transformer~\cite{vaswani2017attention} composed of a three-layer encoder to fit the words and their properties to the saliency values. We combine the word embedding output and the POS embedding vectors together as the input of the transformer, and use the position information again by position coding. The structure of Seq2Saliency is shown in Figure 6.

\subsection{Experiments}
Based on the PrSalM dataset, we train the Seq2Saliency model to fit the mapping relationship between words and their saliency values by taking words and word properties as inputs, and the word saliency values as the output. For the training of the Seq2Saliency model, the sizes of the training set and validation set are 8000 and 1000, respectively.

As shown in Figure 7, we present the change of the loss function values during the training process including the training loss and the validation loss. To give a better view of the change of loss, the loss values in the figure are the sum of the losses of 200 samples. Through the download trend of the training loss, we can see that the Seq2Saliency model is able to converge on the training set. And the download trend of the validation loss shows that the Seq2Saliency model can be used to calculate the saliency values of the text samples that are not appear in the training set. This indicates that it is feasible to calculate the word saliency based on the idea of sequence tagging.

Then we further evaluate the similarity between the results calculated by Seq2Saliency and the results calculated by SHAP. We use the masking saliency (MS)~\cite{li2016understanding} which is also a commonly used saliency calculation method for comparison. In this way,  we take ‘[UNK]’ to replace the words in the original sample in turn, and calculate the confidence difference between the replaced sample and the original sample on the true label to obtain the saliency values of the words. We divide 1000 test samples into 10 groups with 100 samples in each group, and calculate the saliency values of the samples based on MS, Seq2Saliency and SHAP, respectively. In each group, we take the words with saliency values in the top 30\% under different methods and compare the intersection size of the words obtained by MS, Seq2Saliency with the words obtained by SHAP. As shown in Figure 8, the horizontal axis represents the different datasets and different groups, and the vertical axis represents the mean number of overlapping words. It is obvious that the results based on Seq2Saliency are closer to the results based on SHAP than MS. Although it may not reflect that the word saliency calculated by Seq2Saliency is better, it can show that this method is able to realize the mapping between the words and the saliency values.

Finally, we show the saliency heat maps calculated by Seq2Saliency. In Table 3, for each of the three datasets we present three samples. The darker the highlight, the greater the saliency value of the word, that is, the greater the contribution of the word to the prediction result. It can be seen that the saliency heat maps are highly consistent with human intuition. In addition, Seq2Saliency is an end-to-end model, which can directly output the saliency value of each word in the text sample. Compared with most existing black-box saliency calculation methods, Seq2Saliency can obtain saliency values without accessing the original NLP models, and have a high calculation efficiency and a wide range of application scenarios.

\begin{table}[t]\tiny
	\centering
	\begin{tabular}{ ccc }
		\hline
		Dataset & Label & Heat map 
		\\ \hline
		SST-2 & Negative & 
		\begin{minipage}[b]{5cm}
			\centering
			\raisebox{-.4\height}{\includegraphics[width=\linewidth]{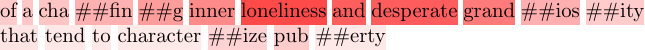}}
		\end{minipage}	
		\\ \hline
		
		SST-2 & Negative & 
		\begin{minipage}[b]{5cm}
			\centering
			\raisebox{-.4\height}{\includegraphics[width=\linewidth]{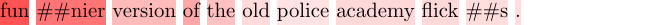}}
		\end{minipage}	
		\\ \hline
		
		SST-2 & Postitive & 
		\begin{minipage}[b]{5cm}
			\centering
			\raisebox{-.4\height}{\includegraphics[width=\linewidth]{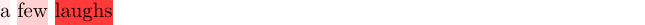}}
		\end{minipage}	
		\\ \hline
		
		AG's News & World & 
		\begin{minipage}[b]{5cm}
			\centering
			\raisebox{-.4\height}{\includegraphics[width=\linewidth]{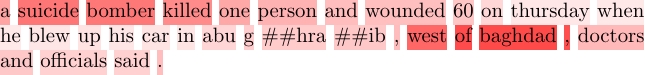}}
		\end{minipage}	
		\\ \hline
		
		AG's News & Sports & 
		\begin{minipage}[b]{5cm}
			\centering
			\raisebox{-.4\height}{\includegraphics[width=\linewidth]{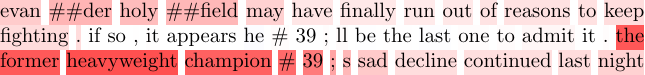}}
		\end{minipage}	
		\\ \hline
		
		AG's News & Business & 
		\begin{minipage}[b]{5cm}
			\centering
			\raisebox{-.4\height}{\includegraphics[width=\linewidth]{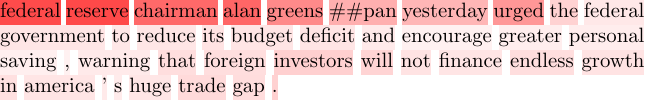}}
		\end{minipage}	
		\\ \hline
		
		SNLI & Entailment & 
		\begin{minipage}[b]{5cm}
			\centering
			\raisebox{-.4\height}{\includegraphics[width=\linewidth]{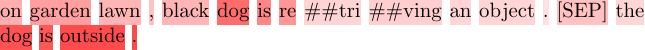}}
		\end{minipage}	
		\\ \hline
		
		SNLI & Contradiction & 
		\begin{minipage}[b]{5cm}
			\centering
			\raisebox{-.4\height}{\includegraphics[width=\linewidth]{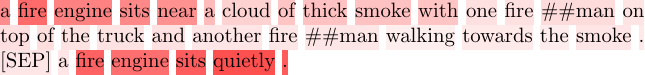}}
		\end{minipage}	
		\\ \hline
		
		SNLI & Neutrality & 
		\begin{minipage}[b]{5cm}
			\centering
			\raisebox{-.4\height}{\includegraphics[width=\linewidth]{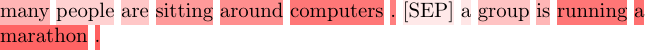}}
		\end{minipage}	
		\\ \hline
	\end{tabular}
	\caption{Examples of saliency heat map calculated by Seq2Saliency}
\end{table}

\section{Conclusion and Future Work}
In this paper, we studied the relationships between the word properties and the saliency values to explain the predictions of NLP models. Based on the analysis, we further established a mapping model, Seq2Saliency, from the words in text samples to the word saliency values based on the idea of sequence tagging. Compared with the previous saliency analysis methods, Seq2Saliency can be used to explain the predictions of the black-box models without needing the internal information such as the gradients. Also, calculating saliency values by Seq2Saliency does not need to access the black-box models, which can greatly improve the saliency calculation efficiency and have a wide range of application scenarios. Based on our experimental data including the word properties and saliency values used in the analysis process and model training, we built PrSalM dataset, which can be used for future NLP explainability research.

Meanwhile, there are many other avenues of future work that we will explore. In addition to classification tasks, we will study Seq2Saliency in other NLP tasks such as machine translation and question answering, in which word properties are also important embodiment of word semantics and word contextual information. 

Besides, the SHAP method is used to get the supervision saliency values in this paper. We will investigate the relationships between the word properties and the saliency based on other saliency analysis methods.

\bibliographystyle{named}
\bibliography{manuscript}

\end{document}